\documentclass[lettersize,journal]{IEEEtran}
\usepackage{amsmath,amsfonts}
\usepackage{algorithmic}
\usepackage{array}
\usepackage[caption=false,font=normalsize,labelfont=sf,textfont=sf]{subfig}
\usepackage{textcomp}
\usepackage{stfloats}
\usepackage{url}
\usepackage{verbatim}
\usepackage{graphicx}
\usepackage{cite}
\usepackage{multirow}
\def\BibTeX{{\rm B\kern-.05em{\sc i\kern-.025em b}\kern-.08em
    T\kern-.1667em\lower.7ex\hbox{E}\kern-.125emX}}
\usepackage{balance}
\begin{document}
\title{Hierarchical Disentangled Representation for Invertible Image Denoising and Beyond}
\author{Wenchao Du, Hu Chen, Yi Zhang \IEEEmembership{Senior Member, IEEE} and Hongyu Yang
\thanks{
W. Du, H. Chen, and H. Yang are with the College of Computer Science, Sichuan University, Chengdu 610065, China. Y. Zhang is with the college of Cyber Science and Engineering, Sichuan University, Chengdu, 610065, China. Email: wenchaodu.scu@gmail.com; huchen@scu.edu.cn; yzhang@scu.edu.cn; yanghongyu@scu.edu.cn.}}

\markboth{}%
{Hierarchical Disentangled Representation for Invertible Image Denoising and Beyond}

\maketitle

\begin{abstract}
  Image denoising is a typical ill-posed problem due to complex degradation. Leading methods based on normalizing flows have tried to solve this problem with an invertible transformation instead of a deterministic mapping. However, the implicit bijective mapping is not explored well. Inspired by a latent observation that noise tends to appear in the high-frequency part of the image, we propose a fully invertible denoising method that injects the idea of disentangled learning into a general invertible neural network to split noise from the high-frequency part. More specifically, we decompose the noisy image into clean low-frequency and hybrid high-frequency parts with an invertible transformation and then disentangle case-specific noise and high-frequency components in the latent space. In this way, denoising is made tractable by inversely merging noiseless low and high-frequency parts. Furthermore, we construct a flexible hierarchical disentangling framework, which aims to decompose most of the low-frequency image information while disentangling noise from the high-frequency part in a coarse-to-fine manner. Extensive experiments on real image denoising, JPEG compressed artifact removal, and medical low-dose CT image restoration have demonstrated that the proposed method achieves competing performance on both quantitative metrics and visual quality, with significantly less computational cost.
\end{abstract}


\section{Introduction}
\IEEEPARstart{I}{mage} denoising, aiming to recover clean observation from its noisy measurement, is a classic inverse problem. Based on the Bayesian perspective, most traditional approaches view it as a general maximum a posteriori (MAP) optimization problem, with assumptions regarding both the image priors and noise (usually Gaussian), which deviate from real cases and lead to critical limitations in practical scenes. Deep learning-based methods \cite{Jain2008NaturalID,Zhang2017BeyondAG,Mao2016ImageRU,Chen2017TrainableNR} have achieved superior denoising performance in recent years. Most of them view it as a nonlinear mapping between noisy and clean image pairs. However, real image noise is generally accumulated from multiple degrading sources, which results in non-injective mapping due to various noise types and levels.

Bridging a bijective transformation between the noisy and clean image pairs to solve such an ambiguous inverse problem in low-level vision has been explored in recent years. Previous methods \cite{Liu2018MultilevelWF,Liu2020DenselySW} have employed convolutional neural networks (CNNs) to model wavelet transforms to solve image restorations, in which wavelet decomposition and reconstruction processes are modeled separately, leading to an injective mapping procedure. Generative adversarial network-based methods \cite{Yue2020DualAN,Du2020LearningIR} have also modeled the bijective transformation for noisy image generation and restoration in supervised and unsupervised  manners. However, these methods need different models to simulate these two processes separately, which leads to complex training procedures.
\begin{figure}[t]
	\centering
	\begin{minipage}[t]{0.24\linewidth}
		\centering
		\includegraphics[width=\textwidth]{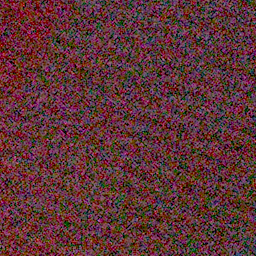}
		\centering{17.55/0.09}
	\end{minipage}
	\begin{minipage}[t]{0.24\linewidth}
		\centering
		\includegraphics[width=\textwidth]{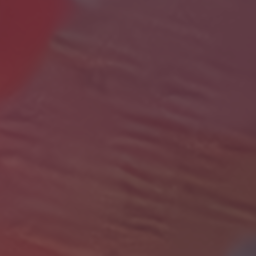}
		\centering{35.30/0.83}
	\end{minipage}
	\begin{minipage}[t]{0.24\linewidth}
		\centering
		\includegraphics[width=\textwidth]{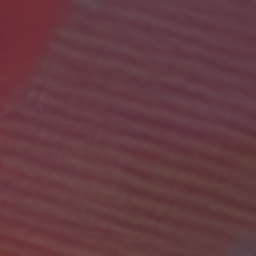}
		\centering{35.56/0.84}
	\end{minipage}
	\begin{minipage}[t]{0.24\linewidth}
		\centering
		\includegraphics[width=\textwidth]{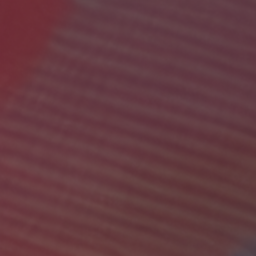}
		\centering{35.92/0.85}
	\end{minipage}
	\begin{minipage}[t]{0.24\linewidth}
		\centering
		\includegraphics[width=\textwidth]{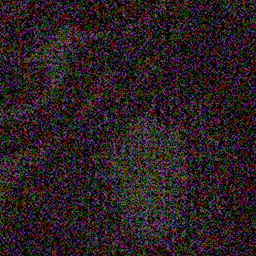}
		\centering{18.47/0.14}
		\centering{Noisy}
	\end{minipage}
	\begin{minipage}[t]{0.24\linewidth}
		\centering
		\includegraphics[width=\textwidth]{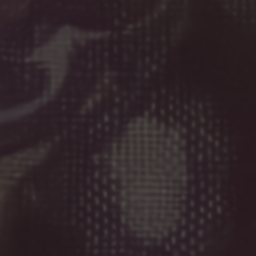}
		\centering{31.62/0.71}
		\centering{InvDn}
	\end{minipage}
	\begin{minipage}[t]{0.24\linewidth}
		\centering
		\includegraphics[width=\textwidth]{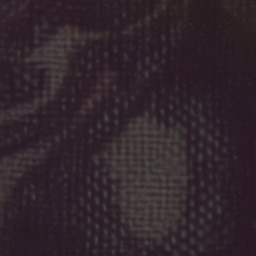}
		\centering{31.91/0.72}
		\centering{DANet}
	\end{minipage}
	\begin{minipage}[t]{0.24\linewidth}
		\centering
		\includegraphics[width=\textwidth]{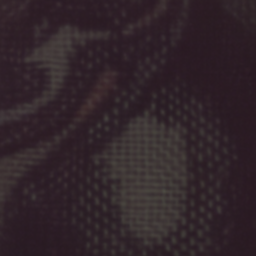}
		\centering{32.25/0.73}
		\centering{\textbf{Ours}}
	\end{minipage}
	\caption{Real image noise removal results on SIDD validation set. The two representative methods are selected for comparisons, i.e. flow-based InvDN \cite{Liu2021InvertibleDN} and GAN-based DANet \cite{Yue2020DualAN}, PSNR and SSIM values are also computed. Zoomed in for better visualization.}
	\label{fig1}
\end{figure}

Normalizing flow that allows for efficient and exact likelihood calculation and sampling by invertible transformation, has been applied to solve ill-posed inverse problems in low-level vision \cite{Abdelhamed2019NoiseFN, Lugmayr2020SRFlowLT, Xiao2020InvertibleIR, Liu2021InvertibleDN}. NoiseFlow \cite{Abdelhamed2019NoiseFN} is a seminal flow-based model for camera noise generation. Unlike general CNN-based methods \cite{Zamir2020CycleISPRI} that model camera imaging pipelines in forward and reverse directions to simulate real noise generation, NoiseFlow learns the distribution of real noise and generates diverse noisy images by latent variable sampling to augment training data. However, it treats the clean image as conditional prior and thus is not fully invertible between clean and noisy image pairs. However, it treats the clean image as conditional prior and thus is not fully invertible. Considering that noise tends to appear in the high-frequency part of the image, InvDn \cite{Liu2021InvertibleDN} discards the high-frequency content, and then utilizes the invertible neural network (INN) \cite{Kingma2018GlowGF} to capture the distribution of the high-frequency part of the clean image, image denoising is achieved by sampling a latent variable from predefined distribution to approximate the lost high-frequency information. Nevertheless, since the model is bijective, the ill-posedness of the task is partly alleviated. However, InvDn assumes that the low and high-frequency contents of the image are independent of each other and thus lacks the ability to exploit their dependency for image denoising, as shown in Fig.~\ref{fig1}, which leads to the recovered image losing part of high-frequency details, e.g., over-smoothed image edges. Furthermore, the implicit bijective mapping is not fully guaranteed due to extra latent variable sampling. 

Instead, in this paper, we propose a fully invertible model for image denoising that injects the idea of disentangled learning into a general invertible architecture to explore feature-level noise-high frequency signals splitting. Specifically, we aim to remove noise from the hybrid high-frequency part of the noisy image while reserving case-specific high-frequency information as much as possible. To do so, we first decompose the image into low and high-frequency representations by an invertible transform during the forward propagation. Assuming the decomposed low-frequency information is noiseless, according to the information lossless characteristics of the normalizing flow, the noise only appears in the high-frequency part. Therefore, our goal is to split the noise component from it, so that the noise-free high-frequency signal is preserved well. 

To this end, we introduce the concept of disentangled learning into the general invertible architecture, which transforms the hybrid high-frequencies into the compact and independent representations with internal characteristics, where we model the high-frequency representation learning in the form of distribution, e.g., the marginal distribution of the hybrid high-frequency representation obeys a pre-specified distribution, i.e. isotropic Gaussian. In this way, we directly disentangle noise and clean high-frequency representations along specific dimensions without extra latent variable sampling, which leads to a more accurate and stable bijective transformation. Image denoising is achieved by inversely merging noise-free low and high-frequency representations only. Furthermore, we hierarchically decompose the high-frequency representation into low and high-frequency parts and disentangle noise from fine-grained high-frequency parts, which results in a flexible and efficient framework for generalized image denoising tasks.

In short, our contributions are summarized as follows: 
\begin{enumerate}{}{}
	\item We propose a fully invertible model for image denoising, which introduces the idea of disentangled learning into a general invertible architecture and achieves a more accurate bijective mapping without latent variable sampling.
	\item We construct a hierarchical high-frequency decomposition framework, which could process diverse noise with varying complexity, and achieve a better trade-off between performance and computational costs.
	\item Extensive experiments on natural image denoising, JPEG compressed artifact removal and medical low-dose CT image restoration demonstrate the proposed method achieves competing performance in terms of quantitative and qualitative evaluations. To the best of our knowledge, this is the first full invertible method capable of solving multiple real image denosing tasks.
\end{enumerate}

The remainder of the paper is organized as follows: Section \ref{related} provides a brief review of related work. Section \ref{Method} presents our approach in detail. In Section \ref{Exps}, extensive experiments are conducted to evaluate the proposed method, and the conclusion is presented in Section \ref{conclusion}.
\section{Related Work}\label{related}
\subsection{Traditional Methods}
\noindent Most traditional image denoising models usually construct a MAP optimization problem with a data filed and an extra regularization term. Along this direction, making assumptions regarding the noise distribution is necessary for most methods \cite{Zoran2011FromLM,Xu2015PatchGB,Chen2015ExternalPP} to build the model (e.g., Mixture of Gaussian). The regularization term is generally based on the natural image prior. Classic total variation \cite{Rudin1992NonlinearTV} uses the statistical characteristics of images to remove noise. Sparse dictionary learning and Field-of-Experts (FoE) also employ certain priors existing in image patches \cite{Dong2011SparsitybasedID,Mairal2009NonlocalSM,Roth2005FieldsOE}. Non-local similarity method \cite{Buades2005ANA} employ non-local similar patterns of image. In addition, transform techniques are also explored, e.g., wavelet domain methods \cite{Kaur2002ImageDU} and block-matching and 3D filtering (BM3D) \cite{Dabov2007ImageDB}. 

\subsection{Deep Learning based Methods}
\noindent Instead of preset image and noise distribution priors, DNNs based methods directly learn a denoiser in a data-driven manner. Previous method \cite{Burger2012ImageDC} first explored the multi-layer perception (MLP). Chen et al. \cite{Chen2017TrainableNR} further proposed a feed-forward deep network called the trainable non-linear reaction diffusion (TNRD) model. Furthermore, Mao et al. \cite{Mao2016ImageRU} utilized a fully convolution encoder-decoder network with symmetric skip connection to solve image restoration. Zhang et al. \cite{Zhang2017BeyondAG} proposed the Gaussian denoising convolution network (DnCNN) and achieved superior performance. Considering non-local self-similarity characteristic of images, non-local attention networks \cite{Liu2018NonLocalRN,Lefkimmiatis2018UniversalDN} are also explored in image restoration.

However, most of these works focus on the synthetic noisy images with specific noise levels, spatially variant noise limit the capacity of such models on real cases. To this end, some methods employed self-adaptive noise level estimated from inputs to serve as extra priors (e.g., FFDNet \cite{Zhang2018FFDNetTA} and CBDNet \cite{Guo2019TowardCB}). VDN \cite{Yue2019VariationalDN} further integrated variational inference into the noise estimation and image denoising with a unique Bayesian framework. AINDNet \cite{Kim2020TransferLF} introduced the transfer learning to mitigate the domain gap between the real and synthetic noise distribution. In addition, simulating real noise with a generative model was also explored in recent works \cite{Yue2020DualAN,Wu2020UnpairedLO}. Considering that DNNs-based image denoising is non-injective in nature, a complex network architecture and powerful representation ability is always required. Thus, Transformer-based methods \cite{Liang2021SwinIRIR,Wang2022UformerAG,Tu2022MAXIMMM} have drawn more attention due to powerful representation learning on global image features.

\subsection{Normalizing Flow based Methods}
\noindent Invertible neural networks (INNs) have drawn more attention to solving ambiguous inverse problems \cite{Ardizzone2018AnalyzingIP}. Unlike DNNs, INNs focus on learning the forward process and using additional latent output variables to capture the information that would otherwise be lost. Due to invertibility, a model of the corresponding inverse process is learned implicitly. NoiseFlow \cite{Abdelhamed2019NoiseFN} modeled the distribution of real noise in ISP imaging process to augment the training data. InvDn \cite{Liu2021InvertibleDN} extended the idea of IRN \cite{Xiao2020InvertibleIR} to real noise removal, which discards hybrid high-frequency information containing noise of the image and exploited extra latent variable sampling to approximate the lost high-frequency information. However, a key factor is ignored that noise is case-specific, which implies the corresponded high-frequency information is also case-specific. More recently, FDN \cite{Liu2021DisentanglingNF} directly disentangled the noise in the latent space with normalizing flows, which leads to a more complex network. FINO \cite{Guo2021FINOFJ} introduced the dual flow models to bridge bijective mapping between the noisy and clean image pairs.

Rather, we aim to solve image denoising by modeling a reliable bijective mapping with single model only. We empirically show that \textit{integrating the idea of disentangled learning into the flow-based framework can result in fast and stable training as well as good performance on generalized image denoising tasks}.

\section{Methodology}\label{Method}
\subsection{Preliminaries}
\noindent Typical INNs models a generative process with a known distribution through a sequence of differentiable, invertible mappings. Formally, let $x_0\in \mathbb{R}^D$ be a random variable with a known and tractable probability density function $p_{\mathcal{X}_0}:\mathbb{R}^D\rightarrow\mathbb{R}$ and let $x_1,\dots,x_N$ be a sequence of random variables such that $x_i=f_i(x_{i-1})$ where $f_i:\mathbb{R}^D\rightarrow \mathbb{R}^D$ is a differentiable, bijective function. Then, if $y=f(x_0)=f_n\circ f_{N-1}\circ \cdots\circ f_1(x_0)$, the change of variables formula says that the probability density function for $y$ is 
\begin{equation}
	p(y)=p_{\mathcal{X}_0}(g(y))\prod_{j=1}^{N}|\det\mathbf{J}_j(g(y))|^{-1}
	\label{eq1}
\end{equation}
where $g=g_1\circ\cdots\circ g_{N-1}\circ g_N$ is the inverse of $f$, and $\mathbf{J}_j=\partial{f_j}/\partial{x}_{j-1}$ is the Jacobian of the $j$th transformation $f_j$ with respect to its input $x_{j-1}$ (\textit{i.e.,} the output of $f_{j-1}$).

Due to such flexibility on accessing to the inverse mapping, INN architecture could be used for variational inference \cite{Kingma2017ImprovedVI,Berg2018SylvesterNF}, and representation learning without any information loss.

INN is composed of basic invertible blocks \cite{Dinh2017DensityEU}. For the $l$-th block, the input $\textbf{u}$ is split into $\textbf{u}_1$ and $\textbf{u}_2$ along the channel axis, and the typical additive affine transformation and corresponded inverse transformation are formulated as:
\begin{align}
	\left\{
	\begin{aligned}
		\textbf{v}_1 &= \textbf{u}_1 + \phi(\textbf{u}_2) \\ 
		\textbf{v}_2 &= \textbf{u}_2 + \eta(\textbf{v}_1)
	\end{aligned}
	\right.
	\Leftrightarrow
	\left\{
	\begin{aligned}
		\textbf{u}_2 &= \textbf{v}_2 - \eta(\textbf{v}_1) \\
		\textbf{u}_1 &= \textbf{v}_1 - \phi(\textbf{u}_2)
	\end{aligned}
	\right.	
	\label{eq2}
\end{align}
where $\phi$ and $\eta$ are arbitrary neural networks. The output of a single block is $[\textbf{v}_1,\textbf{v}_2]$.

To enhance the transformation ability of the identity branch, Eq. (\ref{eq2}) is always augmented as:

\begin{align}
	\left\{
	\begin{aligned}
		\textbf{v}_1 &= \textbf{u}_1 \odot \exp(\psi(\textbf{u}_2)) + \phi(\textbf{u}_2) \\
		\textbf{v}_2 &= \textbf{u}_2 \odot \exp(\rho(\textbf{u}_1)) + \eta(\textbf{u}_1) \\
		\textbf{u}_2 &= (\textbf{v}_2 - \eta(\textbf{v}_1)) \odot \exp(-\rho(\textbf{v}_1)) \\
		\textbf{u}_1 &= (\textbf{v}_1 - \phi(\textbf{u}_2)) \odot \exp(-\psi(\textbf{u}_2))
	\end{aligned}
	\right.
	\label{eq3}
\end{align}
where $\psi(\cdot)$, $\phi(\cdot)$ and $\eta(\cdot)$ denote the transformation functions, which are arbitrary \cite{Dinh2017DensityEU}. Function $\rho(\cdot)$ is further followed by a centered sigmoid function and a scale term to prevent numerical explosion due to the $\exp(\cdot)$ function.

\subsection{Problem Specification}
\noindent We rethink the image denoising task from the perspective of invertible transformation. Suppose the original noisy observation is $y$, its clean image is $x$ and the noise is $n$. We have
\begin{equation}
	p(y)=p(x,n)=p(x)p(n|x)
	\label{eq4}
\end{equation}
where the distribution of noisy image $p(y)$ is a joint distribution corresponded to $x$ and $n$. Directly splitting the noise component from $y$ is difficult due to unknown $p(n)$. In addition, $p(n|x) $ implies the noise is case-specific to image content. A general observation that noise tends to appear in the high-frequency component of the image. Therefore, we employ a wavelet transformation to decompose the noisy image $y$ into low and high-frequency components, denoted as $y_{\text{LF}}$ and $y_{\text{HF}}$ respectively. Then, Eq. (\ref{eq4}) is reformulated by 
\begin{equation}
	p(y)=p(y_{\text{LF}}, {y}_{\text{HF}})=p(y_{\text{LF}})p({y}_{\text{HF}}|y_{\text{LF}})
	\label{eq5}
\end{equation}
Ideally, wavelet decomposition is orthogonal, which could yield dense representation without redundancies, so Eq. (\ref{eq5}) is rewritten as $p(y)=p(y_{\text{LF}}, {y}_{\text{HF}})=p(y_{\text{LF}})p({y}_{\text{HF}})$. 

Our goal is to disentangle noise representation from hybrid high-frequency component, which means that the low-frequency part $y_{\text{LF}}$ is approximately noiseless, and noise is only contained in $p({y}_{\text{HF}})$. Considering the unique characteristics of information lossless in INNs, we could implement it easily. Thus, we focus on how to split noise from $p({y}_{\text{HF}})$ such that
\begin{equation}
	p({y}_{\text{HF}}) = p({y}_{\text{hF}}, y_{\text{n}})=p({y}_{\text{hF}})p(y_{\text{n}}|{y}_{\text{hF}})
	\label{eq6}
\end{equation}
where ${y}_{\text{hF}}$ denotes noise-free high-frequency part, and $y_{\text{n}} $ is case-specific noise.

\begin{figure*}
	\centering
	\includegraphics[width=\textwidth]{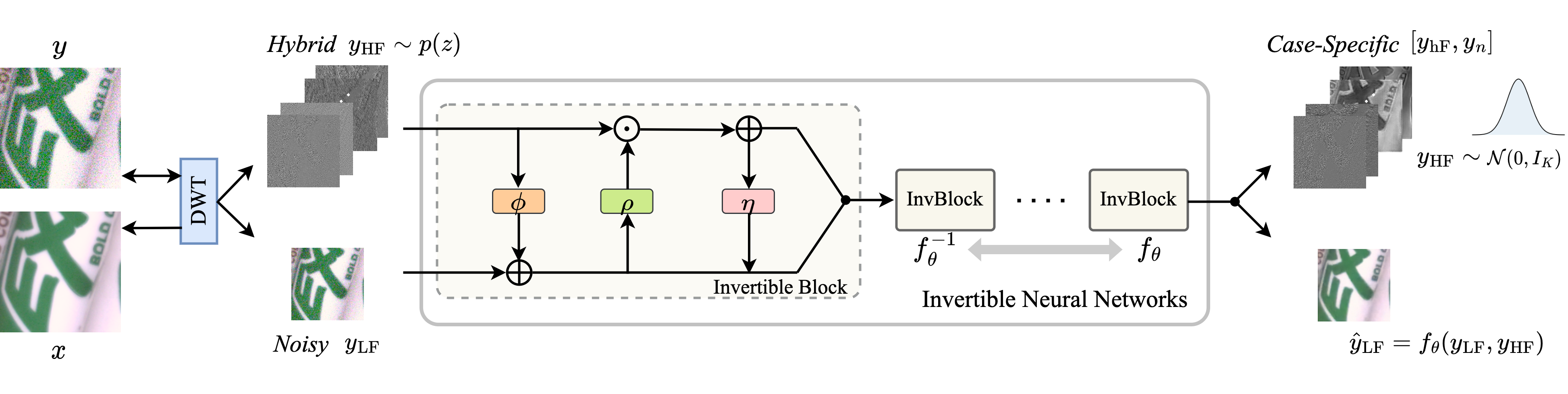}
	\caption{Illustration of invertible image denoising by disentangling representation. In the forward procedure, noisy image $x$ is first decomposed into hybrid high frequencies $y_{\text{HF}}$ and noisy low frequencies $y_{\text{LF}}$ with Discrete Wavelet Transform (DWT), which are feed into a parameterized invertible neural network $f_{\theta}(\cdot)$ to transform into a clean low-frequency $\hat{y}_{\text{LF}}$ and a high-frequency $y_{\text{HF}}$ obeying specific distribution in the latent space, where case-specific high-frequency $y_{\text{hF}}$ and noise $y_{\text{n}}$ would be disentangled. Image denoising is achieved through the inverse function $f^{-1}_{\theta}$ and DWT operation with disentangled $y_{\text{hF}}$ and clean $\hat{y}_{\text{LF}}$.}
	\label{fig2}
\end{figure*}

Different from general flow based models, e.g., IRN \cite{Xiao2020InvertibleIR} and InvDn \cite{Liu2021InvertibleDN},  a latent variable $z$ is introduced to approximate the case-specific ${y}_{\text{hF}}$, and transformed $z$ to be case-agnostic with a specified distribution, where $z$ is used to model the high-frequency information lost in degrading. For real image denoising, $y_{\text{n}}$ and ${y}_{\text{hF}}$ are always case-specific and tangled. Therefore, it is difficult to model lost ${y}_{\text{hF}}$ accurately with a case-agnostic variable without any conditional priors. Thus, we attempt to solve this problem with a simple manner, i.e. disentangling noise part from ${y}_{\text{HF}}$ in the latent space. In this way, case-specific ${y}_{\text{hF}}$ and $y_{\text{n}}$ can be split well.

Directly disentangling hybrid ${y}_{\text{HF}}$ is difficult due to unknown $p(y_{\text{n}})$. To this end, we assume that ${y}_{\text{hF}}$ and ${y}_{\text{n}}$ are independent, so that Eq. (\ref{eq6}) is reformulated as
\begin{equation}
	p({y}_{\text{HF}}) = p({y}_{\text{hF}}, y_{\text{n}})=p({y}_{\text{hF}})p(y_{\text{n}})
	\label{eq7}
\end{equation}

To satisfy this assumption, we directly enforce $p({y}_{\text{HF}}) $ to obey pre-specific distribution, e.g., isotropic Gaussian, which means $p({y}_{\text{HF}}) $ is compact and explainable, a key characteristic is that the feature vectors in ${y}_{\text{HF}}$ are independent of each other. We disentangle the high-frequency representation ${y}_{\text{HF}}$ along the specific dimension to split ${y}_{\text{hF}}$ and ${y}_{\text{n}}$. To reconstruct clean observation, we explicitly remove ${y}_{\text{n}}$ and only preserve specific $p({y}_{\text{hF}}) $, so that $p({y}_{\text{HF}})= p({y}_{\text{hF}})p(y_{\text{n}}) = p({y}_{\text{hF}}) \cdot 1$. Benefited from the invertible characteristic of INNs, image denoising could be achieved by inverse pass using Eq. (\ref{eq1}).

\subsection{Model Architecture}
\noindent The sketch of disentangling framework is illustrated in Fig.~\ref{fig2}, which contains two processes: i.e. \textit{forward decomposition} and \textit{inverse reconstruction}, and is easily injected into the general invertible framework.

\subsubsection{Forward Decomposition}
The key to our approach is to decompose the approximately noiseless low-frequency and case-specific high-frequency representations during the forward pass. To achieve this, we first exploit the Discrete Wavelet Transform (DWT) to decompose the image into low and high-frequency parts, where Haar Transformation \cite{Lienhart2002AnES} is used to approximate it for simplicity. Haar Transform explicitly decomposes the inputs, (i.e. images or a group of feature maps) into an approximate low-pass representation and high-frequency coefficients with three directions. More concretely, it transforms the input with shape $(H\times W \times C)$ into a tensor of shape $(\frac{1}{2}H \times \frac{1}{2}W \times 4C)$. The first $C$ slices of the output tensor are effectively produced by average pooling, which is approximately a low-pass representation equivalent to the bilinear interpolation down-sampling. The remaining three groups of $C$ slices contain residual components in the vertical, horizontal and diagonal directions, which are the high-frequency information of the original input. By such a transformation, the low and high-frequency information are explicitly separated.

Then, an invertible network module is used to further abstract the ${y}_{\text{LF}}$ and ${y}_{\text{HF}}$, where we leverage the coupling layer architecture in \cite{Dinh2015NICENI,Dinh2017DensityEU}, presented in i.e. Eqs. (\ref{eq2}) and (\ref{eq3}). Our goal is to polish the low and high-frequency inputs to obtain a suitable low-frequency representation and an independent properly distributed high-frequency representation. Therefore, we match ${y}_{\text{LF}}$ and ${y}_{\text{HF}}$ respectively to the split of $\textbf{u}_1$, $\textbf{u}_2$ in Eq. (\ref{eq2}). Furthermore, to increase the model capacity, we employ the additive transformation (Eq. (\ref{eq2})) for the low-frequency part $\textbf{u}_1$, and the enhanced affine transformation (Eq. (\ref{eq3})) for the high-frequency part $\textbf{u}_2$.

After transformation, a noise-free low-frequency representation $\hat{y}_{\text{LF}}$ is expected. However, noise also appears in $\hat{y}_{\text{LF}}$ actually due to non-ideal decomposition. Inspired by the Nyquist-Shannon sampling theorem that the lost information during down-sampling a clean image amounts to high-frequency contents, we utilize the bicubic method \cite{Mitchell1988ReconstructionFI} to guide $\hat{y}_{\text{LF}}$ decomposition. Let $x_{guide}$ be the down-sampled clean image $x$ corresponding to $\hat{y}_{\text{LF}}$ that is produced by the bicubic method. To generate the clean low-frequency representation, we drive the $\hat{y}_{\text{LF}}$ to resemble ${x}_{guide}$:
\begin{equation}
	L_{guide}:=\ell_{\mathcal{X}}(x_{guide},\hat{y}_{\text{LF}})
	\label{eq8}
\end{equation}  
where $\ell_{\mathcal{X}}$ is a difference metric on $\mathcal{X}$, i.e. the $L_2$ loss. Benefited from the characteristic of information lossless of invertible architecture, our low-frequency guidance loss drives noise appearing in ${y}_{\text{HF}}$ only.

Another goal during the forward pass is to disentangle noise from high-frequency part ${y}_{\text{HF}}$. To this end, we enforce the ${y}_{\text{HF}}$ to obey specific distribution, i.e. $p_{{y}_{\text{HF}}}\sim\mathcal{N}(\textbf{0},I_K)$ so that it could be disentangled along the specific dimension. A KL divergence loss is used as the distribution metric, where
\begin{equation}
D_{KL}=-\int{p(z)\log(\frac{p(z)}{q(z)})\text{d} z}
\label{eq9}
\end{equation}
which is referred to as our distribution guidance loss $L_{dist}$.

\subsubsection{Inverse Reconstruction}
Considering feature vectors in transformed ${y}_{\text{HF}}$ is independent, we directly split it along the channel axis into a specific high-frequency part ${y}_{\text{hF}}$ and noise ${y}_{\text{n}}$, i.e. ${y}_{\text{HF}}=[{y}_{\text{hF}}, {y}_{\text{n}}]$, and ${y}_{\text{hF}}$ and ${y}_{\text{n}}$ are case-specific. To reconstruct the clean image $x$ in the flow-based architecture, we need to construct dimension-consistent noise-free high-frequency representation $\hat{y}_{\text{HF}}$. To do so, we simply replace ${y}_{\text{n}}$ with $\textbf{0}$ in reconstructed high-frequency representation $\hat{y}_{\text{HF}}$, i.e. $\hat{y}_{\text{HF}}=[{y}_{\text{hF}}, \textbf{0}]$, which lies in a subspace of ${y}_{\text{HF}}$. Further, to guide noise component disentanglement, we minimize reconstruction loss $L_{recon}$ with corresponded clean image $x$:
\begin{equation}
	L_{recon}:=\ell_{\mathcal{X}}(x,\text{iDWT}(f^{-1}_{\theta}(\hat{y}_{\text{LF}},\hat{y}_{\text{HF}})))
	\label{eq10}
\end{equation}
where $\ell_{\mathcal{X}}$ measures the difference between the clean image and the reconstructed one, i.e. the $L1$ loss. $\text{iDWT}$ denotes inverse DWT, and $f^{-1}_{\theta}$ denotes the inverse pass of parameterized invertible neural networks.

In addition, implicit noisy image self-reconstruction could also be achieved by injecting case-specific ${y}_{\text{n}}$ into ${y}_{\text{HF}}$, it doesn't rely on any extra self-supervised constraints. This implies there exists an implicit bijective mapping between the noisy and clean image pairs.
\begin{figure*}
	\centering
	\includegraphics[width=\textwidth]{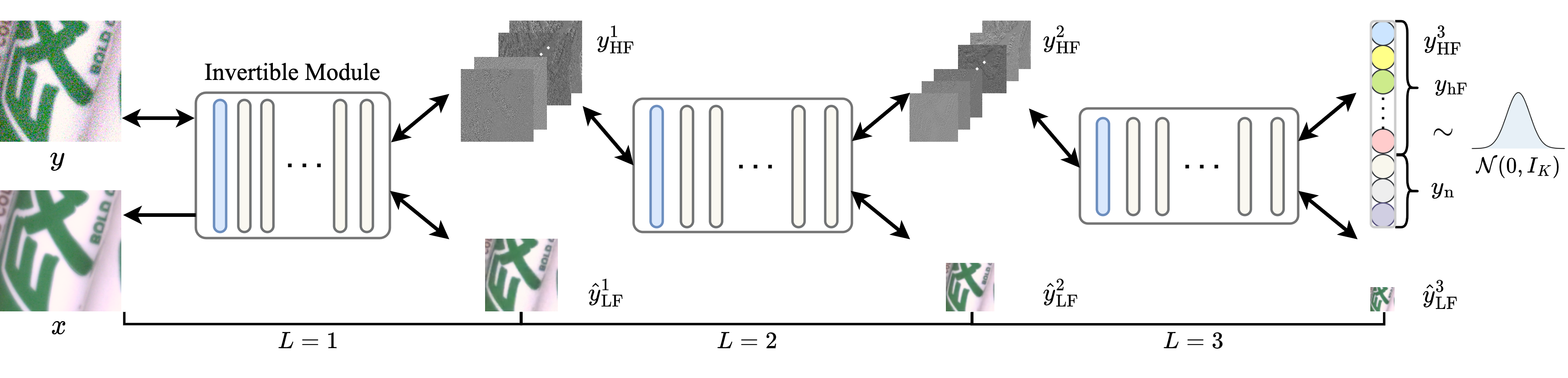}
	\caption{Illustration of our hierarchical disentangling framework. Our approach exploits the three invertible modules to transform the noisy image $y$ into clean low frequencies and specific high frequencies with different scales, i.e. $L=3$. Each invertible module is composed of a DWT layer and stacked InvBlocks.}
	\label{fig3}
\end{figure*}

\subsection{Hierarchical Disentangled Representation}
Furthermore, inspired by the latent observation that most image information is located in the low-frequency part, and high and low-frequencies are not decomposed exactly due to non-ideal transform. Thus, exploiting the sufficient low-frequency information while disentangling fine-grained high-frequency signals is necessary. Based on this analysis, we construct a multi-level decomposing framework in Fig.~\ref{fig3}.

During the forward propagating, we stack multiple invertible modules to decompose the high-frequency representation ${y}_{\text{HF}}$ into low and high-frequency parts, this leads to multiscale low-frequency outputs $\{\hat{y}^l_{\text{LF}}\}^{L}_{l=1}$. Therefore, the low-frequency guidance loss $L_{guide}$ is rewritten as
\begin{equation}
	L_{guide}:=\sum_{l=1}^{L}\ell_{\mathcal{X}}({x}^{l}_{guide},\hat{y}^{l}_{\text{LF}})
	\label{eq11}
\end{equation}
where $L$ is the number of levels decomposed, $\hat{y}^{l}_{\text{LF}}$ represents the decomposed low-frequency output in $l$th level, and corresponded multiscale guided image $x^l_{guide}$ is generated by down-sampling $x$ to ${1}/{2^l}$ scale with the bicubic method. 

For the high-frequency part, we only minimize the KL divergence loss for the last output $y^L_{\text{HF}}$, and disentangle noise component from it. In our experiments, we set $L=3$ at most for generalized image denoising tasks.

In inverse propagation, we reconstruct clean high frequencies $\hat{y}^l_{\text{HF}}$ level by level, from level $L$ to level $1$. For $l$th level, it leads to an implicit conditional flow model, i.e. $p(\hat{y}^{l-1}_{\text{HF}}|{y}^{l}_{\text{HF}}, \hat{y}^l_{\text{LF}})$, which could be implemented by $\hat{y}^{l-1}_{\text{HF}}=f^{-1}_{\theta}({y}^{l}_{\text{HF}}, \hat{y}^l_{\text{LF}})$. Therefore, the case-specific high-frequency signal is generated in a coarse-to-fine manner, it is efficient and stable. 

We optimize the whole architecture by minimizing the compact loss $L_{total}$ with the combination of reconstruction loss $L_{recon}$, low-frequency guidance loss $L_{guide}$ and distribution loss $L_{dist}$:
\begin{equation}
	L_{total}:=\lambda_{1}L_{recon}+\lambda_2L_{guide}+\lambda_3L_{dist}
	\label{eq12}
\end{equation}
where $\lambda_1,\lambda_2$ and $\lambda_3$ are coefficients for balancing different loss terms.

\begin{figure}
	\centering
	\begin{minipage}[t]{0.49\linewidth}
		\centering
		\includegraphics[width=\textwidth]{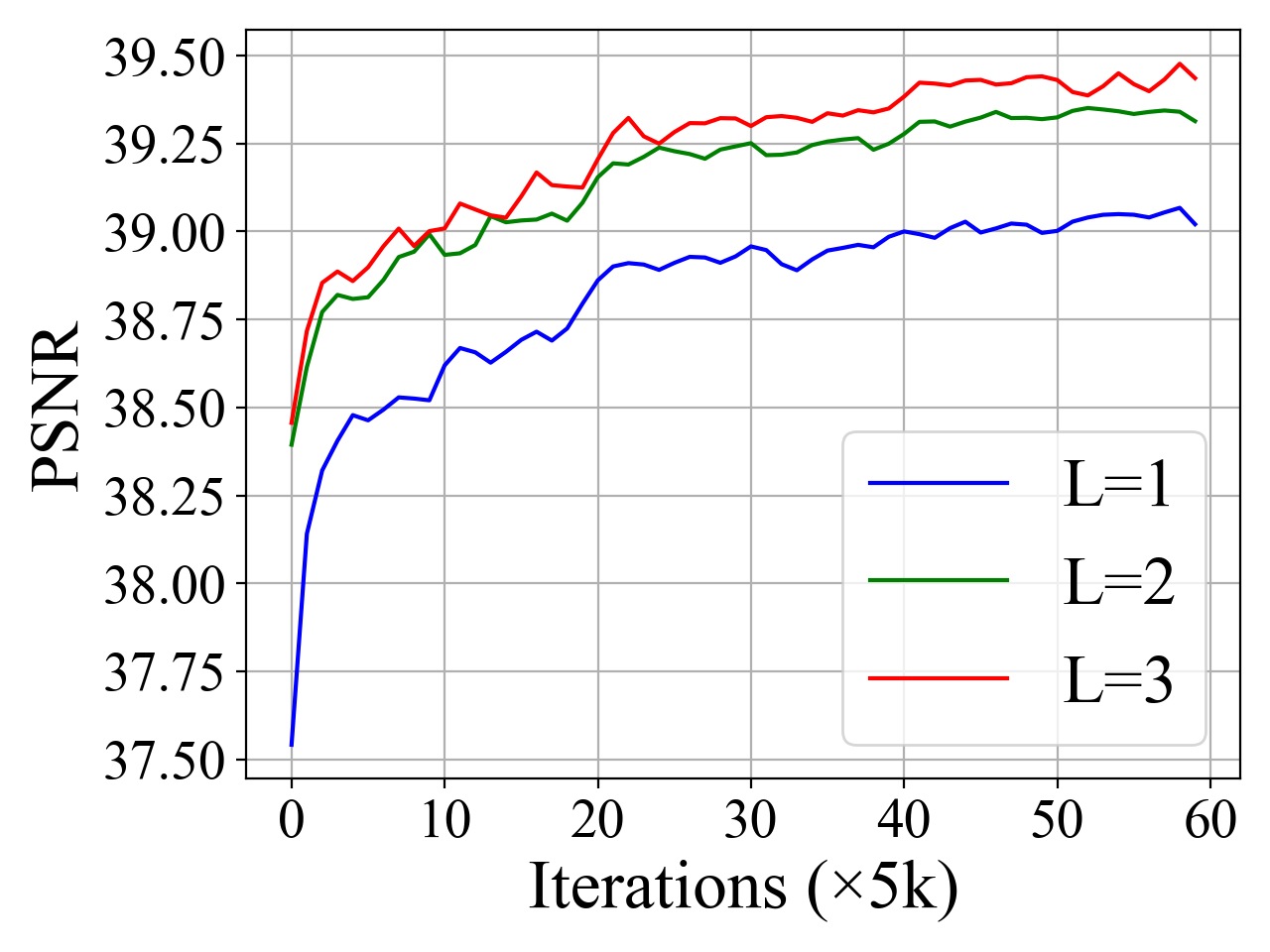}
		\centerline{$(\text{a})$}
	\end{minipage}
	\begin{minipage}[t]{0.49\linewidth}
		\centering
		\includegraphics[width=\textwidth]{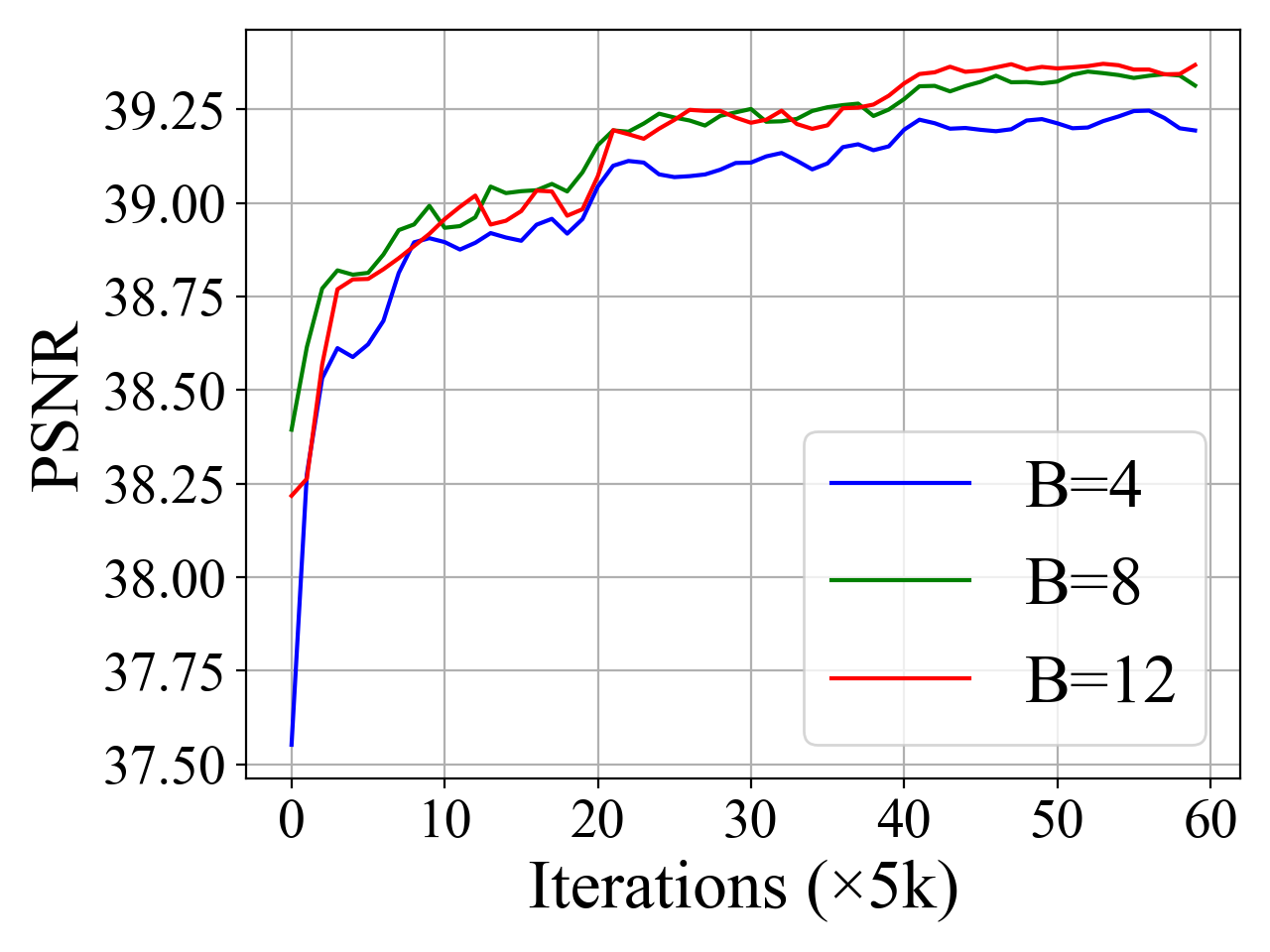}
		\centerline{$(\text{b})$}
	\end{minipage}
	\caption{Online PSNR curves on SIDD validation set with $300k$ iterations.}
	\label{fig4}
\end{figure}
\begin{table}
  \centering
	\caption{The architecture configurations.}
  \label{table1}
  \setlength{\tabcolsep}{4.4mm}{
	\begin{tabular}{l|c|c|c|c}
		\hline
		\multirow{2}{*}{Metric} & \multicolumn{2}{c}{Single-level} & \multicolumn{2}{c}{Multi-level (Ours)} \\
		\cline{2-3}
		\cline{4-5}
		& L=2 & L=3 & L=2 & L=3 \\
		\hline
		\it{PSNR} & 39.358 & \textbf{39.499} & \textbf{39.364} & 39.468 \\
		\it{SSIM} & 0.908 & 0.910 & \textbf{0.909} & 0.910 \\
		\it{\#Param(M)} & 4.419 & 12.554 & \textbf{4.021} & \textbf{9.092} \\
		\hline
	\end{tabular}
  }
\end{table}
\begin{table}[t]
  \centering
	\caption{The disentangled dimension configurations for model with B=8 and L=2.}
	\label{table2}
  \setlength{\tabcolsep}{4.6mm}{
	\begin{tabular}{l|c|c|c|c}
		\hline
		DIM$(y_\text{n})$  & 4/5 & 3/5 & 2/5 & 1/5 \\
		\hline
		\it{PSNR} & 39.348 & 39.352 & \textbf{39.403} & 39.364 \\
		\it{SSIM} & 0.9086 & 0.9085 & 0.9086 & 0.9086 \\
		\hline
	\end{tabular}
  }
\end{table}

\section{Experiment}\label{Exps}
\subsection{Experimental Setting}
\subsubsection{Datasets}
To validate the effectiveness of our method, two representative real image noise datasets, the Smartphone Image Denoising Dataset (SIDD) \cite{Abdelhamed2018AHD} and Darmstadt Noise Dataset (DND) \cite{Pltz2017BenchmarkingDA}, are utilized to verify our method's performance. The SIDD is taken by five smartphone cameras with small apertures and sensor sizes from 10 scenes under varied lighting conditions. Ground truth images are generated through a systematic procedure. We use the medium version of SIDD as the training set, which contains 320 clean-noisy pairs for training and 1280 cropped patches from the other 40 pairs for validation. The reported test results are obtained via an online submission system. The DND is captured by four consumer-grade cameras of different sensor sizes. It contains 50 pairs of real-world noisy and approximately noise-free images. These images are cropped into 1000 patches of size $512\times512$. Similarly, the performance is evaluated by submitting the results to the online system. Considering DND does not provide any training data, we employ a training strategy by combining the training set of SIDD and Renoir \cite{Anaya2018RENOIRA}. Results are submitted to the DND benchmark by utilizing the same model that provides the best validation performance on the SIDD benchmark. 

\begin{figure*}
	\centering
	\begin{minipage}[t]{0.16\linewidth}
		\centering
		\includegraphics[width=\textwidth]{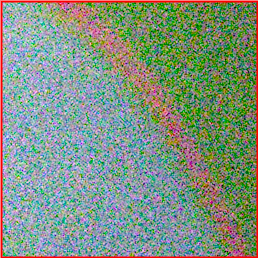}
		\centerline{$(\text{a})$}
	\end{minipage}
	\begin{minipage}[t]{0.16\linewidth}
		\centering
		\includegraphics[width=\textwidth]{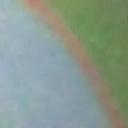}
		\centerline{$(\text{b})$}
	\end{minipage}
	\begin{minipage}[t]{0.16\linewidth}
		\centering
		\includegraphics[width=\textwidth]{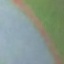}
		\centerline{$(\text{c})$}
	\end{minipage}
	\begin{minipage}[t]{0.16\linewidth}
		\centering
		\includegraphics[width=\textwidth]{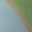}
		\centerline{$(\text{d})$}
	\end{minipage}
	\begin{minipage}[t]{0.16\linewidth}
		\centering
		\includegraphics[width=\textwidth]{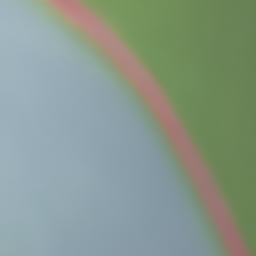}
		\centerline{$(\text{e})$}
	\end{minipage}
	\begin{minipage}[t]{0.16\linewidth}
		\centering
		\includegraphics[width=\textwidth]{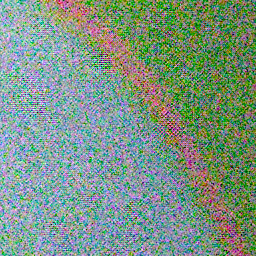}
		\centerline{$(\text{f})$}
	\end{minipage}
  \subfloat{
	\begin{minipage}[t]{0.104\linewidth}
		\centering
		\includegraphics[width=\textwidth]{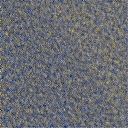}
	\end{minipage}
	\begin{minipage}[t]{0.104\linewidth}
		\centering
		\includegraphics[width=\textwidth]{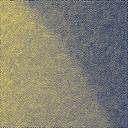}
	\end{minipage}
	\begin{minipage}[t]{0.104\linewidth}
		\centering
		\includegraphics[width=\textwidth]{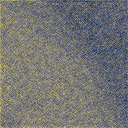}
	\end{minipage}
	\begin{minipage}[t]{0.104\linewidth}
		\centering
		\includegraphics[width=\textwidth]{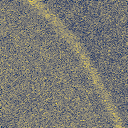}
	\end{minipage}
	\begin{minipage}[t]{0.104\linewidth}
		\centering
		\includegraphics[width=\textwidth]{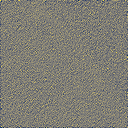}
	\end{minipage}
	\begin{minipage}[t]{0.104\linewidth}
		\centering
		\includegraphics[width=\textwidth]{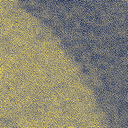}
	\end{minipage}
	\begin{minipage}[t]{0.104\linewidth}
		\centering
		\includegraphics[width=\textwidth]{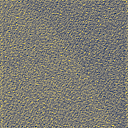}
	\end{minipage}
	\begin{minipage}[t]{0.104\linewidth}
		\centering
		\includegraphics[width=\textwidth]{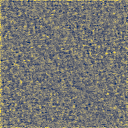}
	\end{minipage}
	\begin{minipage}[t]{0.104\linewidth}
		\centering
		\includegraphics[width=\textwidth]{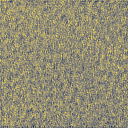}
	\end{minipage}
  }\\
  \subfloat{
	\begin{minipage}[t]{0.104\linewidth}
		\centering
		\includegraphics[width=\textwidth]{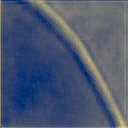}
	\end{minipage}
	\begin{minipage}[t]{0.104\linewidth}
		\centering
		\includegraphics[width=\textwidth]{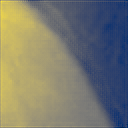}
	\end{minipage}
	\begin{minipage}[t]{0.104\linewidth}
		\centering
		\includegraphics[width=\textwidth]{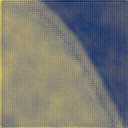}
	\end{minipage}
	\begin{minipage}[t]{0.104\linewidth}
		\centering
		\includegraphics[width=\textwidth]{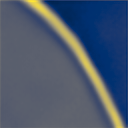}
	\end{minipage}
	\begin{minipage}[t]{0.104\linewidth}
		\centering
		\includegraphics[width=\textwidth]{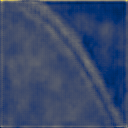}
	\end{minipage}
	\begin{minipage}[t]{0.104\linewidth}
		\centering
		\includegraphics[width=\textwidth]{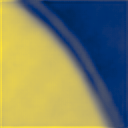}
	\end{minipage}
	\begin{minipage}[t]{0.104\linewidth}
		\centering
		\includegraphics[width=\textwidth]{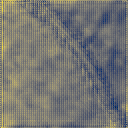}
	\end{minipage}
	\begin{minipage}[t]{0.104\linewidth}
		\centering
		\includegraphics[width=\textwidth]{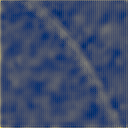}
	\end{minipage}
	\begin{minipage}[t]{0.104\linewidth}
		\centering
		\includegraphics[width=\textwidth]{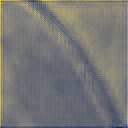}
	\end{minipage}
  }
	\caption{Visualization analysis for invertible bijective mapping. The top row denotes the results of invertible image denoising. (a) is input noisy image, (b), (c), and (d) denote the decomposed low-frequency outputs from the level $L=1,2,3$, separately. (e) is inverse denoised output, and (f) is the self-reconstructed noisy image. The second row visualizes the high-frequency feature maps from $L=1$ during forward propagating. The bottom row visualizes the disentangled high-frequency feature maps from $L=1$ during inverse propagating.}
	\label{fig5}
\end{figure*}

\subsubsection{Implementations}
The proposed method cascades three invertible modules at most, each module contains a Haar transform layer and 8 basic invertible blocks, i.e. $L=3, B=8$.  For the single invBlock (as shown in Fig.~\ref{fig2}), we implement the non-linear functions $\phi$, $\rho$, and $\eta$ with the Densenet Block (DB) \cite{Wang2018ESRGANES}. All the models are trained with Adam as the optimizer, with momentum of $\beta_1=0.9,\beta_2=0.999$. The batch size is set as 16 with $144\times144$ size, and the initial learning rate is fixed at $2\times10^{-4}$, which decays by half for every $100k$ iterations. The training is performed on a single Tesla P100 GPU. We augment the training data with extra horizontal and vertical flipping, as well as random rotations. For loss functions, we set $\lambda_1=1,\lambda_2=4$ and $\lambda_3=1$ separately for different loss terms in all experiments. In addition, Peak-Signal-Noise-Ratio (PNSR) and Structural Similarity (SSIM) are used to evaluate the performance of methods in all experiments.

\subsection{Ablation Study}
\noindent We mainly explore three major determinants of our model: a). Model capacity, which depends on the number of decomposed levels, and the number of invertible blocks in single invertible module; b). Effectiveness of architecture, including decomposing types of image and disentangling ways of framework; and c). Disentangled representation. All the experiments are performed in SIDD validation set with $300k$ iterations.

\subsubsection{Model capability}
We first study two key factors affecting the model capability, i.e. multi-level decomposition and the capacity of the single invertible module. As shown in Fig.~\ref{fig4}-(a) and (b), we observe that increasing the decomposed levels leads to significant performance gains. In addition, fixing decomposition levels (e.g., $L=2$), stacking more invBlocks in the single invertible module also further enhances the ability of the model, but it also brings more parameters. Therefore, we set the $\text{B}=8$ and $L=3$ at most in our architecture configurations. 


\subsubsection{Architecture Designing}
We also consider different architectures, i.e. decompose low and high-frequency parts in the last level only instead of each level, which is similar to InvDn \cite{Liu2021InvertibleDN}. The quantitative results are illuminated in Tab.~\ref{table1}, our multilevel decomposition architecture with high-frequencies disentangling achieves a better trade-off in terms of the performance and complexity of the model.

\subsubsection{Disentangled Representation}
We further explore the effects of disentangled dimensions in $y_\text{HF}$. We split $y_{\text{n}}$ from $y_{\text{HF}}$ with different dimensions along the channel axis. All models are trained separately. Tab.~\ref{table2} gives the detailed results, the best model is achieved by setting $\text{DIM}(y_{\text{n}}) = 2/5 \cdot \text{DIM}(y_{\text{HF}})$. We use it in our final model configurations.

\subsubsection{Analysis for Bijective Mapping}
We further explore the bijective relationship in our architecture. As shown in Fig.~\ref{fig5}, we first give the decomposed low-frequency outputs from multi-level decomposition during forward propagation, denoised output, and self-reconstructed noisy input by implicitly inverse reconstruction. Our method only splits case-specific noise from the hybrid high-frequency component in the latent space to achieve image denoising, where it bridges the bijective transformation between the noisy image generation and restoration. Moreover, we visualize the decomposed hybrid high-frequency components and disentangled noise-free high-frequencies components. It is obvious that our method could remove case-specific noise signals in latent space while retaining fine high-frequencies, which implies the effectiveness of our method.

\begin{figure*}[t]
	\centering
	\begin{minipage}[t]{0.16\linewidth}
		\centering
		\includegraphics[width=\textwidth]{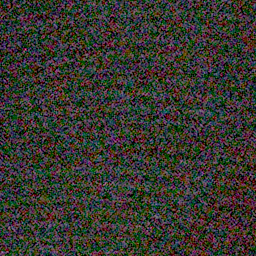}
		\centerline{(17.56/0.1110)} \\
		\includegraphics[width=\textwidth]{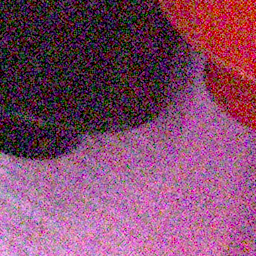}
		\centerline{(18.17/0.1205)}
		\centerline{Noisy}
	\end{minipage}
	\begin{minipage}[t]{0.16\linewidth}
		\centering
		\includegraphics[width=\textwidth]{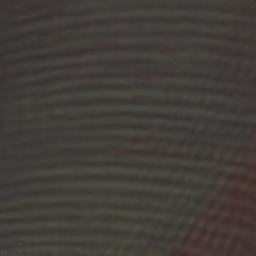}
		\centerline{(34.19/0.7965)}
		\includegraphics[width=\textwidth]{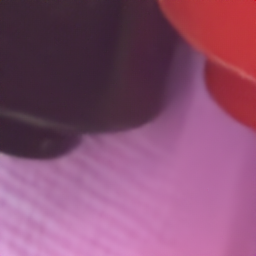}
		\centerline{(35.08/0.8442)}
		\centerline{DANet \cite{Yue2020DualAN}}
	\end{minipage}
	\begin{minipage}[t]{0.16\linewidth}
		\centering
		\includegraphics[width=\textwidth]{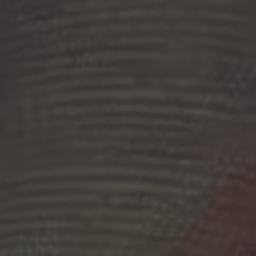}
		\centerline{(34.00/0.7857)}
		\includegraphics[width=\textwidth]{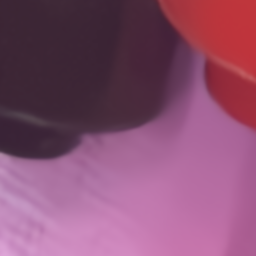}
		\centerline{(34.74/0.8365)}
		\centerline{InvDn \cite{Liu2021InvertibleDN}}
	\end{minipage}
	\begin{minipage}[t]{0.16\linewidth}
		\centering
		\includegraphics[width=\textwidth]{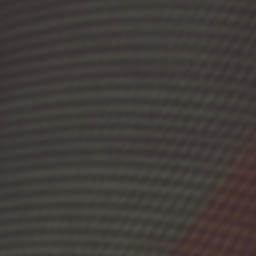}
		\centerline{(34.80/0.8156)}
		\includegraphics[width=\textwidth]{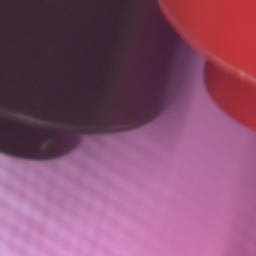}
		\centerline{(35.68/0.8564)}
		\centerline{MAXIM \cite{Tu2022MAXIMMM}}
	\end{minipage}
	\begin{minipage}[t]{0.16\linewidth}
		\centering
		\includegraphics[width=\textwidth]{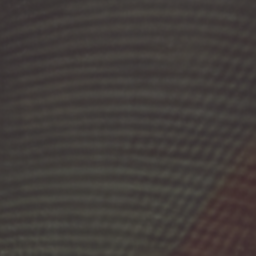}
		\centerline{(34.47/0.8067)}
		\includegraphics[width=\textwidth]{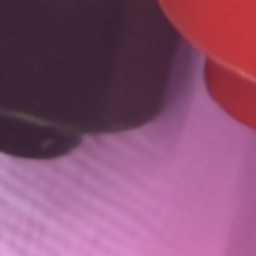}
		\centerline{(35.14/0.8458)}
		\centerline{\textit{\textbf{Ours}(L=2)}}
	\end{minipage}
	\begin{minipage}[t]{0.16\linewidth}
		\centering
		\includegraphics[width=\textwidth]{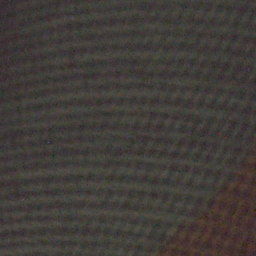}
		\centerline{(PSNR/SSIM)}
		\includegraphics[width=\textwidth]{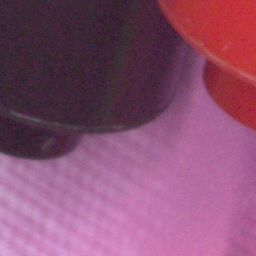}
		\centerline{(PSNR/SSIM)}
		\centerline{Reference}
	\end{minipage}
	\begin{minipage}[t]{0.16\linewidth}
		\centering
		\includegraphics[width=\textwidth]{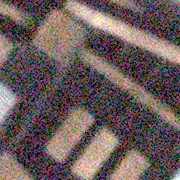}
		\centerline{Noisy}
		\centerline{(PSNR/SSIM)}
	\end{minipage}
	\begin{minipage}[t]{0.16\linewidth}
		\centering
		\includegraphics[width=\textwidth]{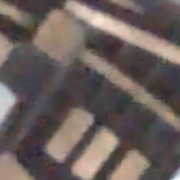}
		\centerline{CBDNet \cite{Guo2019TowardCB}}
		\centerline{(31.40/0.8364)}
	\end{minipage}
	\begin{minipage}[t]{0.16\linewidth}
		\centering
		\includegraphics[width=\textwidth]{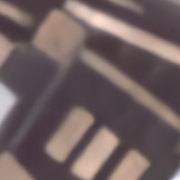}
		\centerline{VDN \cite{Yue2019VariationalDN}}
		\centerline{(34.08/0.9166)}
	\end{minipage}
	\begin{minipage}[t]{0.16\linewidth}
		\centering
		\includegraphics[width=\textwidth]{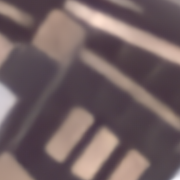}
		\centerline{DANet \cite{Yue2020DualAN}}
		\centerline{(33.66/0.9148)}
	\end{minipage}
	\begin{minipage}[t]{0.16\linewidth}
		\centering
		\includegraphics[width=\textwidth]{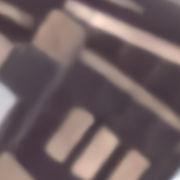}
		\centerline{InvDn \cite{Liu2021InvertibleDN}}
		\centerline{(34.22/0.9216)}
	\end{minipage}
	\begin{minipage}[t]{0.16\linewidth}
		\centering
		\includegraphics[width=\textwidth]{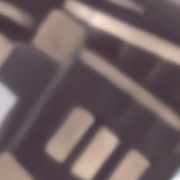}
		\centerline{\textit{\textbf{Ours}(L=3)}}
		\centerline{(\textbf{34.28/0.9224})}
	\end{minipage}
	\caption{Visualized denoising samples from SIDD and DND datasets. The first and second rows are from the SIDD, and bottom is from the DND.}
	\label{fig6}
\end{figure*}
\begin{table}[t]
	\caption{Comprehensive comparisons with other competing methods.}
	\label{table3}
  \setlength{\tabcolsep}{1.9mm}{
	\begin{tabular}{l|c|c|c|c|c|c}
		\hline
		\multirow{2}{*}{Method} & \multirow{2}{*}{\shortstack{\it{\#Param}\\\it{(M)}}} & \multirow{2}{*}{\shortstack{\it{MACs}\\\it{(G)}}} & \multicolumn{2}{c}{SIDD \cite{Abdelhamed2018AHD}} & \multicolumn{2}{c}{DND \cite{Pltz2017BenchmarkingDA}}\\
		\cline{4-5}
		\cline{6-7}
		& & & \it{PSNR} & \it{SSIM} & \it{PSNR} & \it{SSIM} \\
		\hline
		DnCNN \cite{Zhang2017BeyondAG} & 0.67 & 44.02 & 37.73 & 0.941 & 37.90 & 0.9430  \\ 
		TNRD \cite{Chen2017TrainableNR}  & -- & -- & 24.73 & 0.643 & 33.65 & 0.8306  \\
		BM3D \cite{Dabov2007ImageDB} & -- & -- & 25.65 & 0.685 & 33.51 & 0.8507  \\
		CBDNet \cite{Guo2019TowardCB} & 4.34 & -- & 33.28 & 0.868 & 38.06 & 0.9421  \\
		RIDNet \cite{Anwar2019RealID} & 1.50 & 98.26  & -- & -- & 39.26 & 0.8306 \\
		GradNet \cite{Liu2020GradNetID} & 1.60 & --  & 38.34 & 0.946 & 39.44 & 0.9543 \\
		AINDNet \cite{Kim2020TransferLF} & 13.76 & -- & 39.08 & 0.953 & 39.53 & 0.9561  \\
		VDN \cite{Yue2019VariationalDN} & 7.82 & 49.5 & 39.28 & 0.909 & 39.38 & 0.9518  \\
		MIRNet \cite{Zamir2020LearningEF} &31.78 & 816.0  & \textbf{39.72} & 0.959 & \textbf{39.88} & 0.9563\\
		SADNet \cite{Chang2020SpatialAdaptiveNF} & 4.23 & 18.97 & -- & -- & 39.59 & 0.9523 \\
		MPRNet \cite{Zamir2021MultiStagePI} &15.7 & 1394.0  & \textbf{39.71} & 0.958 & \textbf{39.80} & 0.9540\\
		MAXIM \cite{Tu2022MAXIMMM} & 22.20 & 339.0 & \textbf{39.96} & 0.960 & \textbf{39.84} & 0.9540 \\
		\hline
		DANet \cite{Yue2020DualAN} & 63.01 & 14.85 & 39.25 & 0.955 & 39.47 & 0.9548  \\
		DANet++ \cite{Yue2020DualAN} & 63.01 & 14.85 & 39.43 & 0.956 & 39.58 & 0.9545  \\
		InvDn \cite{Liu2021InvertibleDN} & 2.64 & 47.96 & 39.28 & 0.955 & 39.57 & 0.9522 \\ 
		FDN \cite{Liu2021DisentanglingNF} & 4.38 & 77.64 & 39.31 & 0.955 & -- & -- \\
		FINO \cite{Guo2021FINOFJ} & -- & -- & 39.40 & 0.957 & -- & -- \\
		\hline
		Ours+RB+L2 & 2.47 & 46.54 & \textbf{39.31}& 0.956 & -- & --\\
		Ours+RB+L3  & 5.26 & 52.26 & \textbf{39.40} & 0.957& -- & -- \\
		Ours+DB+L2 & 4.02 & 74.02 & \textbf{39.39}& 0.956& \textbf{39.54}& 0.9520\\
		Ours+DB+L3 & 9.09 & 84.40 & \textbf{39.48} & 0.957& \textbf{39.60}& 0.9536\\
		\hline
	\end{tabular}
  }
\end{table}
\subsection{Real Image Noise Removal}
\noindent We perform comprehensive comparisons, including model parameters, computational complexity (MACs), and quantitative metrics, on two real denoising benchmarks, i.e. SIDD \cite{Abdelhamed2018AHD} and DND \cite{Pltz2017BenchmarkingDA}. Note that ``\textit{MACs}" is counted on a single RGB image with $256\times256$ size. Moreover, we select two kinds of representative methods for comparisons, one is the deterministic mapping-based methods, including CNN-based and Transformer methods; The other is bijective mapping-based methods, e.g., Flow-based and GAN-based methods. In addition, in order to demonstrate the effectiveness of our method, we further explore the lightweight architecture configurations, e.g., using the more lightweight invBlocks with Residual Block (RB) \cite{Lim2017EnhancedDR} and different decomposition levels.

Tab.~\ref{table3} lists the detailed results, there exists a latent tendency for the deterministic mapping-based methods that the larger models bring better performance, e.g., MIRNet and MPRNet, but which also lead to expensive computational costs. Moreover, MAXIM introduces the MLP-style Transformer achieve significant performance gains on the SIDD test set, but it also brings great computations. Instead, bijective mapping-based methods reverse this tendency, e.g., DANet and InvDn. They achieve the better trade-off between the performance and computational complexity. 

Our method further balances the performance and computational costs on the SIDD test set, where we only stack two lightweight invertible modules (denote as ``RB+L2'') and achieves better results compared to general CNN-based methods and bijective mapping-based methods. Meanwhile, the performance could be improved further by extending to 3-level decomposition (denote as ``RB+L3''). Compared with DANet and InvDn, the PSNRs of our lightweight models increase by $0.1\sim 0.2$dB on the SIDD test set. Replacing lightweight residual blocks with dense blocks (DB) in a single InvBlock, our method (denote as ``DB+L2'' and ``DB+L3'') achieves consistent performance gains on two real denoising benchmarks. In addition, our method is more flexible and friendly to mobile applications, which achieves comparable results with state-of-the-art methods, e.g., MIRNet, MPRNet, and MAXIM, note that the performance of our approach could be further improved with the more effective inverse architecture, e.g., invertible attention networks \cite{Sukthanker2022GenerativeFW}, but it is beyond the scope of this paper.

\begin{table*}[t]
	\caption{PSNR$\vert$ SSIM $\vert$ \textbf{PSNR-B} values comparisons. The best and the second best results are boldfaced and \underline{underlined}.}
	\label{table4}
  \setlength{\tabcolsep}{1.3mm}{
		\begin{tabular}{ccccc|ccc|ccc|ccc|ccc|ccc}
			\hline
			Dataset & QF & \multicolumn{3}{c|}{SADCT \cite{Foi2007PointwiseSD}} &   \multicolumn{3}{c|}{LD \cite{Li2014ACE}} &   \multicolumn{3}{c|}{PCA \cite{Song2020CompressedIR}} &   \multicolumn{3}{c|}{ARCNN \cite{Dong2015CompressionAR}} &  \multicolumn{3}{c|}{TNRD \cite{Chen2017TrainableNR}} &  \multicolumn{3}{c}{DnCNN \cite{Zhang2017BeyondAG}} \\
			\hline
			
			\multirow{3}{*}{\textit{Classic5}} & 10 & 28.88  & 0.807  & 28.16 & 28.39 & 0.780 & 27.59 & 29.32 & 0.800 & 29.08 & 29.03 & 0.793 & 28.76 & 29.28 & 0.799 & 29.04 & 29.40 & 0.803 & 29.13 \\
			& 20 & 30.92 & 0.866 & 29.75 & 30.30 & 0.858 & 29.37 & 31.56 & 0.858 & 31.12 & 31.15 & 0.852 & 30.59 & 31.47 & 0.858 & 31.05 & 31.63 & 0.861 & 31.19 \\
			& 30 & 32.14 & 0.891 & 30.83 & 31.47 & 0.833 & 30.17 & 32.86 & 0.884 & 32.31 & 32.51 & 0.881 & 31.98 & 32.78 & 0.884 & 32.24 & 32.91 & 0.886 & 32.38 \\
			\hline
			\multirow{3}{*}{\textit{LIVE1}} & 10 & 28.65 & 0.809 & 28.01 & 28.26 & 0.805 & 27.68 & 29.01 & 0.809 & 28.83 & 28.96 & 0.808 & 28.77 & 29.14 & 0.811 & 28.88 & 29.19 & 0.812 & 28.90 \\
			& 20 & 30.81 & 0.878 & 29.82 & 30.19 & 0.872 & 29.64 & 31.28 & 0.875 & 30.72 & 31.29 & 0.873 & 30.79 & 31.46 & 0.877 & 31.04 & 31.59 & 0.880 & 31.07 \\
			& 30 & 32.08 & 0.908 & 30.92 & 29.41 & 0.896 & 29.36 & 32.62 & 0.903 & 32.18 & 32.67 & 0.904 & 32.22 & 32.84 & 0.906 & 32.28 & 32.98 & 0.909 & 32.34 \\  
			\hline
			\multirow{3}{*}{\textit{BSD500}} & 10 & 28.23 & 0.778 & 27.38 & 28.03 & 0.782 & 27.29 & 28.64 & 0.779 & 28.01 & 28.56 & 0.791 & 27.87 & 28.60 & 0.793 & 27.95 & 28.84 & 0.801 & 28.44 \\
			& 20 & 30.09 & 0.851 & 28.61 & 29.82 & 0.851 & 28.43 & 30.73 & 0.851 & 29.42 & 30.43 & 0.859 & 29.10 & 30.51 & 0.861 & 29.34 & 31.05 & 0.874 & 30.29 \\
			& 30 & 31.21 & 0.884 & 29.34 & 30.87 & 0.872 & 29.15 & 31.99 & 0.884 & 30.84 & 31.52 & 0.890 & 29.92 & 31.58 & 0.890 & 30.02 & 32.36 & 0.905 & 31.43 \\  
			\hline
			\textit{Twitter} & & 27.61 & 0.728 & 27.53 & 27.58 & 0.727 & 27.49 & 27.71 & 0.730 & 27.66 & 27.54 & 0.730 & 27.49 & 27.60 & 0.727 & 27.52 & 27.63 & 0.729 & 27.54 \\
			\hline
			\textit{WeChat} & & 29.60 & 0.780 & 29.59 & 29.48 & 0.796 & 29.47 & 29.63 & 0.799 & 29.60 & 29.30 & 0.799 & 29.29 & 26.64 & 0.791 & 26.63 & 29.57 & 0.798 & 29.57 \\
			\hline
			\hline 
			Dataset & QF & \multicolumn{3}{c|}{LIPIO \cite{Fan2021AGD}} &   \multicolumn{3}{c|}{M-Net \cite{Tai2017MemNetAP}} &   \multicolumn{3}{c|}{DCSC \cite{Fu2019JPEGAR}} &   \multicolumn{3}{c|}{RNAN \cite{Zhang2019ResidualNA}} &  \multicolumn{3}{c|}{\textit{\textbf{Ours}(L=2)}} &  \multicolumn{3}{c}{\textit{\textbf{Ours}(L=3)}} \\
			\hline
			\multirow{3}{*}{\textit{Classic5}} & 10 & 29.35  & 0.802  & 29.04 & 29.69 & 0.811 & 29.31 & \underline{29.62} & \underline{0.827} & 29.30 & \textbf{29.87} & \textbf{0.828} & 29.42 & 29.51 &  0.816& \underline{29.47} & 29.53 & 0.815 & \textbf{29.50} \\
			& 20 & 31.58 & 0.857 & 31.12 & \textbf{31.90} & 0.866 & 31.29 & \underline{31.81} & \textbf{0.880} & 31.34 & 32.11 & 0.869 & 32.16 & 31.57 & 0.870 & \underline{31.52} & 31.64 & \underline{0.871} & \textbf{31.59} \\
			& 30 & 32.86 & 0.884 & 32.28 & 32.97 & 0.888 & \underline{32.49} & \underline{33.06} & \textbf{0.903} & 32.49 & \textbf{33.38} & 0.892 & 32.35 & 32.70 & 0.894 & \underline{32.62} & 32.81 & \underline{0.894} & \textbf{32.72} \\
			\hline
			\multirow{3}{*}{\textit{LIVE1}} & 10 & 29.17 & 0.812 & 28.89 & \underline{29.45} & 0.819 & 29.04 & 29.34 & \textbf{0.832} & 29.01 & \textbf{29.63} & \underline{0.824} & 29.13 & 29.36 & 0.824 & \underline{29.32} & 29.37 & 0.823 & \textbf{29.33} \\
			& 20 & 31.52 & 0.877 & 31.07 & \underline{31.83} & 0.885 & 31.14 & 31.70 & \textbf{0.896} & 31.18 & \textbf{32.03} & 0.888 & 31.12 & 31.64 & 0.889 & \textbf{31.57} & 31.65 & \underline{0.889} & \textbf{31.57} \\
			& 30 & 32.99 & 0.907 & 32.31 & \textbf{33.07} & 0.911 & 32.47 & \textbf{33.07} & \textbf{0.922} & 32.43 & 33.45 & 0.915 & 32.22 & 32.94 & 0.915 & \textbf{32.83} & 32.95 & \underline{0.916} & \textbf{32.83} \\  
			\hline
			\multirow{3}{*}{\textit{BSD500}} & 10 & 28.81 & 0.782 & 28.39 & \underline{28.96} & 0.804 & 28.56 & 28.95 & \underline{0.805} & 28.55 & \textbf{29.08} & \textbf{0.805} & 28.48 & 28.97 & 0.796 & \underline{28.92} & 28.97 & 0.795 & \textbf{28.93} \\
			& 20 & 30.92 & 0.855 & 30.07 & 31.05 & 0.874 & 30.36 & \underline{31.13} & \textbf{0.876} & 30.41 & \textbf{31.25} & \underline{0.875} & 30.27 & 31.10 & 0.868 & \underline{31.01} & 31.11 & 0.868 & \textbf{31.02} \\
			& 30 & 32.31 & 0.887 & 31.27 & \underline{32.61} & 0.907 & 31.15 & 32.42 & \underline{0.906} & 31.52 & \textbf{32.70} & \textbf{0.907} & 31.33 & 32.34 & 0.899 & \underline{32.22} & 32.35 & 0.899 & \textbf{32.22} \\  
			\hline
			\textit{Twitter} & & 27.47 & 0.733 & 27.41 & 27.98 & 0.744 & 27.87 & 27.63 & 0.731 & 27.43 & 27.43 & 0.718 & 27.42 & \underline{31.04} & \underline{0.794} & \underline{30.82} & \textbf{31.09} & \textbf{0.795} & \textbf{30.90}  \\
			\hline
			\textit{WeChat} & & 28.90 & 0.800 & 28.90 & 29.82 & 0.807 & 29.82 & 29.58 & 0.800 & 29.58 & 29.56 & 0.800 & 29.56 & \textbf{32.32} & \underline{0.823} & \textbf{32.07} & \underline{32.30} & \textbf{0.823} & \textbf{32.07}  \\
			\hline
		\end{tabular}
  }
\end{table*}
\begin{figure*}[t]
	\centering
	\subfloat{
		\begin{minipage}[b]{0.258\linewidth}
			\centering
			\includegraphics[width=\textwidth]{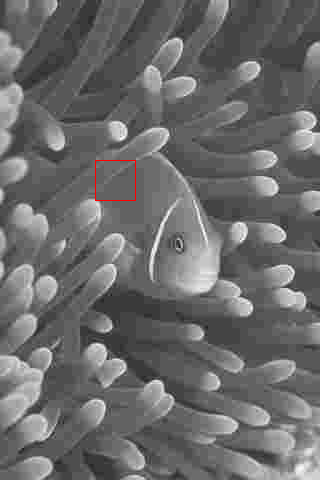}
			\centerline{\textit{Example in ``BSD500''}}
			\centerline{(PSNR-B/SSIM)}
		\end{minipage}
	}
	\centering
	\subfloat{
		\begin{minipage}[b]{0.1681\linewidth}
			\centering
			\includegraphics[width=\textwidth]{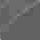}
			\centerline{JPEG}
			\centerline{(29.43/0.8595)}
			\includegraphics[width=\textwidth]{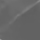}
			\centerline{DnCNN \cite{Zhang2017BeyondAG}}
			\centerline{(33.67/0.9223)}
		\end{minipage}
	}
	\centering
	\subfloat{
		\begin{minipage}[b]{0.1681\linewidth}
			\centering
			\includegraphics[width=\textwidth]{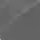}
			\centerline{LD \cite{Li2014ACE}}
			\centerline{(31.34/0.8963)}
			\includegraphics[width=\textwidth]{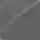}
			\centerline{LIPIO \cite{Fan2021AGD}}
			\centerline{(32.80/0.9139)}
		\end{minipage}
	}
	\subfloat{
		\begin{minipage}[b]{0.1681\linewidth}
			\centering
			\includegraphics[width=\textwidth]{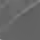}
			\centerline{PCA \cite{Song2020CompressedIR}}
			\centerline{(32.90/0.9173)}
			\includegraphics[width=\textwidth]{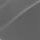}
			\centerline{DCSC \cite{Fu2019JPEGAR}}
			\centerline{(33.55/0.9314)}
		\end{minipage}
	}
	\subfloat{
		\begin{minipage}[b]{0.1681\linewidth}
			\centering
			\includegraphics[width=\textwidth]{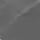}
			\centerline{ARCNN \cite{Dong2015CompressionAR}}
			\centerline{(33.12/0.9148)}
			\includegraphics[width=\textwidth]{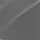}
			\centerline{\textit{\textbf{Ours}(L=2)}}
			\centerline{(\textbf{34.12/0.9327})}
		\end{minipage}
	}
	\caption{Visualized comparisons on synthetic datasets with JPEG QF=10. Red rectangle denotes the zoomed ROI area. }
	\label{fig7}
\end{figure*}
Visualized comparisons are shown in Fig.~\ref{fig6}. It is clear that other bijective mapping-based methods (e.g., DANet and InvDn) could remove noise well but also bring over-smoothed effects, e.g., blurred edges and local structures. Transformer-based MAXIM could alleviate this problem by capturing non-local high-frequency patterns with global self-attention. Instead, our method only removes the noise appearing in the high-frequency textures of degraded images, which doesn't depend on any nonlocal patterns or image priors, while with finer details against others'.

\begin{figure*}[t]
	\centering
	\subfloat{
		\begin{minipage}[b]{0.413\linewidth}
			\centering
			\includegraphics[width=\textwidth]{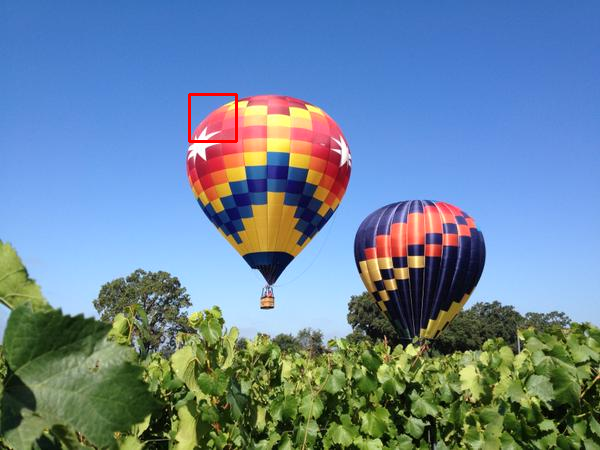}
			\centerline{\textit{Example in ``Twitter"}}
			\centerline{ }
		\end{minipage}
	}
	\centering
	\subfloat{
		\begin{minipage}[b]{0.13\linewidth}
			\centering
			\includegraphics[width=\textwidth]{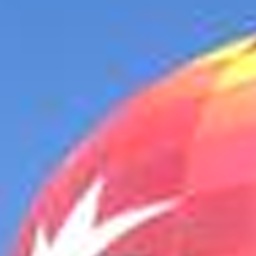}
			\centerline{\textit{Compressed}}
			\centerline{(25.57/0.7280)}
			
			\includegraphics[width=\textwidth]{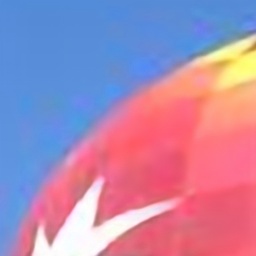}
			\centerline{DnCNN \cite{Zhang2017BeyondAG}}
			\centerline{(26.35/0.7534)}
		\end{minipage}
	}
	\centering
	\subfloat{
		\begin{minipage}[b]{0.13\linewidth}
			\centering
			\includegraphics[width=\textwidth]{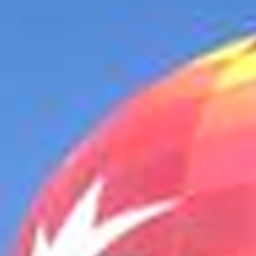}
			\centerline{PCA \cite{Song2020CompressedIR}}
			\centerline{(25.60/0.7319)}
			\includegraphics[width=\textwidth]{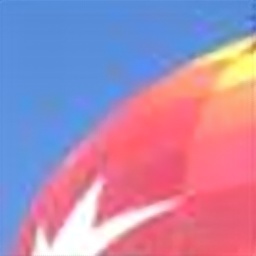}
			\centerline{LIPIO \cite{Fan2021AGD}}
			\centerline{(25.71/0.7337)}
		\end{minipage}
	}
	\centering
	\subfloat{
		\begin{minipage}[b]{0.13\linewidth}
			\centering
			\includegraphics[width=\textwidth]{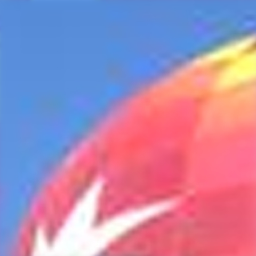}
			\centerline{LD \cite{Li2014ACE}}
			\centerline{(25.59/0.7322)}
			\includegraphics[width=\textwidth]{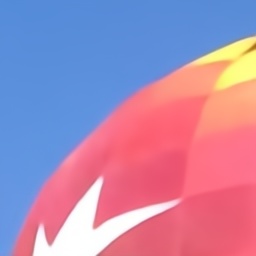}
			\centerline{\textit{\textbf{Ours}(L=2)}}
			\centerline{(\textbf{31.27/0.8375})}
		\end{minipage}
	}
	\centering
	\subfloat{
		\begin{minipage}[b]{0.13\linewidth}
			\centering
			\includegraphics[width=\textwidth]{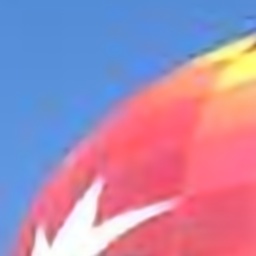}
			\centerline{ARCNN \cite{Dong2015CompressionAR}}
			\centerline{(26.24/0.7488)}
			\includegraphics[width=\textwidth]{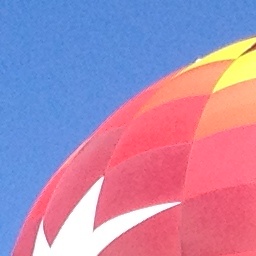}
			\centerline{Reference}
			\centerline{(PSNR-B/SSIM)}
		\end{minipage}
	}
	\caption{Visualized comparisons on the real-world use case. Red rectangle denotes the zoomed ROI area.}
	\label{fig8}
\end{figure*}
\subsection{JPEG Compression Artifact Removal}
\noindent Beyond real image denoising, our method is further extended to JPEG compression artifact reduction. Following the same experimental setting as in \cite{FuLearningDP}, we first validate our approach in synthetic datasets, where we use both the training and testing sets from BSD500 \cite{Arbelez2011ContourDA} as training data. JPEG compressed images were generated by the Matlab JPEG encoder. To present the performance of our method on blind image deblocking, the JPEG quality factors (QF) are randomly generated in $[1,40]$. \textit{Note that we only train one single model (i.e. blind image compressed artifact removal) to handle all the JPEG compression factors, e.g., 10, 20 and 30}. The whole training process was conducted on the Y channel image of YCrCb space. 

In contrast to synthetic JPEG deblocking, real image compressed artifact reduction generally contains two implicit tasks, i.e., up-scaling and artifact reduction, where the original images are usually compressed and rescaled for transmission and storage. Two real datasets, which are collected from popular social medias but with different compression rates, are used to verify our method's effectiveness, i.e., \textit{Twitter} \cite{Dong2015CompressionAR} and \textit{WeChat} \cite{FuLearningDP}. \textit{Twitter} contains 114 training images and extra 10 images for validation. Each high-resolution image ($3264\times 2448$) results in a compressed and rescaled version with a fixed resolution of $600\times 450$. For \textit{WeChat}, which only provides 300 testing images, each image ($3000 \times 4000$ pixel) has a corresponded compression version with $600 \times 800$. Our approach is only trained on \textit{Twitter} training set, and tested in validation set and \textit{WeChat}.

\subsubsection{Comparisons on synthetic datasets}
Tab.~\ref{table4} lists the detailed results on the three widely used synthetic datasets, i.e., 5 images in \textit{Classic5} \cite{Zeyde2010OnSI}, 29 images in \textit{LIVE1}\cite{Sheikh2005LIVE} and 100 images in the validation set of \textit{BSD500}, where we select some representative traditional methods \cite{Foi2007PointwiseSD,Li2014ACE,Song2020CompressedIR} and deep learning based methods \cite{Dong2015CompressionAR,Zhang2017BeyondAG,Chen2017TrainableNR,Tai2017MemNetAP,Fu2019JPEGAR,Zhang2019ResidualNA} for comparison. In addition, we use the PSNR, SSIM and the PSNR-B \cite{Yim2011QualityAO} for quantitative evaluations, where PSNR-B is more sensitive to blocking artifacts than the PSNR.

As shown in Tab.~\ref{table4}, although the proposed method doesn't achieve the best PSNR and SSIM metrics, it has the best PSNR-B values against other methods, which implies that the proposed method is more effective on recovering the local high-frequencies of the compressed image than other methods. Further, we observe a latent tendency that our method achieves approximate consistent results in terms of the PSNR and PSNR-B metrics. In contrast, other methods all exhibit obvious degradation for the single PSNR-B metric.

Visualized results are demonstrated in Fig.~\ref{fig7}, the obvious blocking effects couldn't be reduced well in local high-frequency areas with general CNN-based methods. Instead, our approach can recover consistent textures and smoother edges while reducing blocking artifacts significantly.

\subsubsection{Comparisons on real cases} 
To avoid out-of-memory caused by excessive image resolution during inference, we first crop the image with a sliding window, which results in local image blocks without overlap, and then perform artifact reduction operation and measurement calculation.

Quantitative results are demonstrated in Tab.~\ref{table4}. Traditional and CNN-based methods all appear heavy performance degradation due to complex noise distribution and high-frequency information loss. Instead, our approach still achieves obvious performance gains on \textit{Twitter} and \textit{WeChat}, i.e. the PSNR and PSNR-B increase by $2\sim 3$ dB on average, which implies that our method is more effective to process real compressed artifact removal. Furthermore, visualized results are illuminated in Fig.~\ref{fig8}, our method presents significant advantages in removing real artifacts while preserving fine high-frequency details.

\begin{figure*}
	\centering
		\begin{minipage}[t]{0.245\linewidth}
			\centering
			\includegraphics[width=\textwidth]{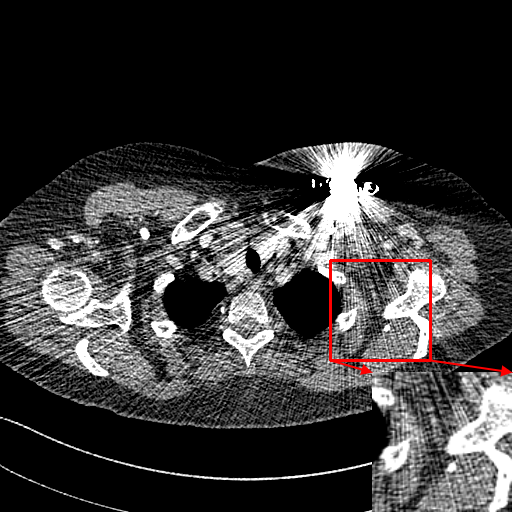}
			\centerline{LDCT}
			\centerline{(34.39/0.9318)}
      \includegraphics[width=\textwidth]{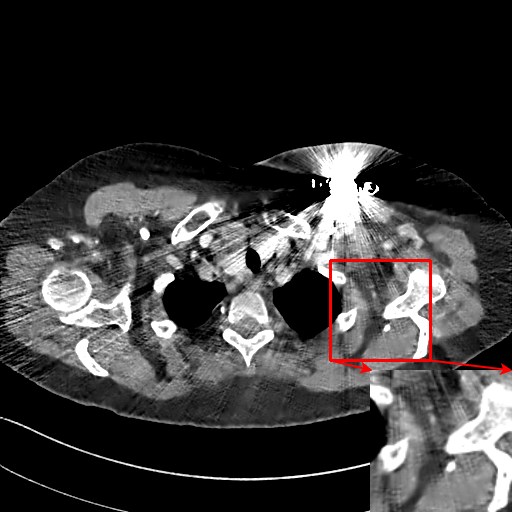}
			\centerline{CTformer \cite{Wang2022CTformerCT}}
			\centerline{(37.80/0.9758)}
			
		\end{minipage}
    \begin{minipage}[t]{0.245\linewidth}
			\centering
			\includegraphics[width=\textwidth]{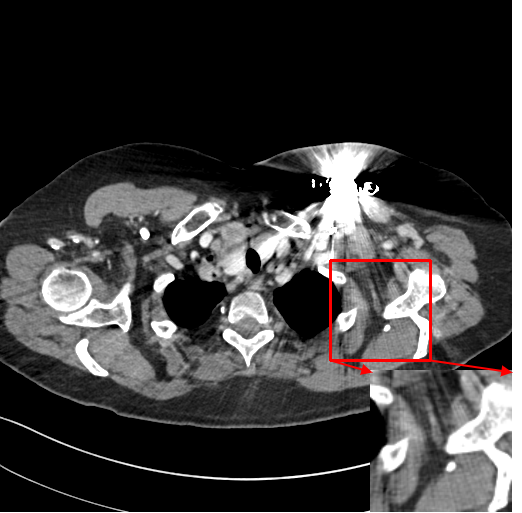}
			\centerline{RedCNN \cite{Chen2017LowDoseCW}}
			\centerline{(38.84/0.9813)}
			\includegraphics[width=\textwidth]{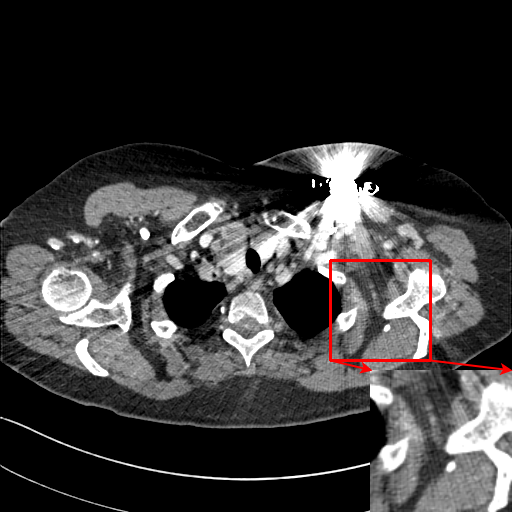}
			\centerline{Eformer \cite{Luthra2021EformerEE}}
			\centerline{(39.21/0.9799)} 
		\end{minipage}
		\begin{minipage}[t]{0.245\linewidth}
			\centering
      \includegraphics[width=\textwidth]{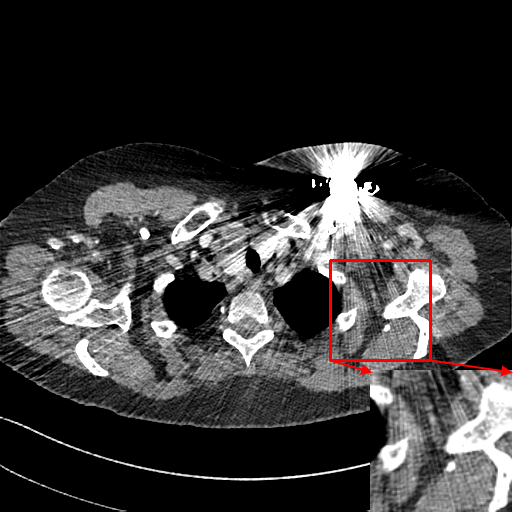}
			\centerline{WGAN-VGG \cite{Yang2018LowDoseCI}}
			\centerline{(35.98/0.9598)}
			\includegraphics[width=\textwidth]{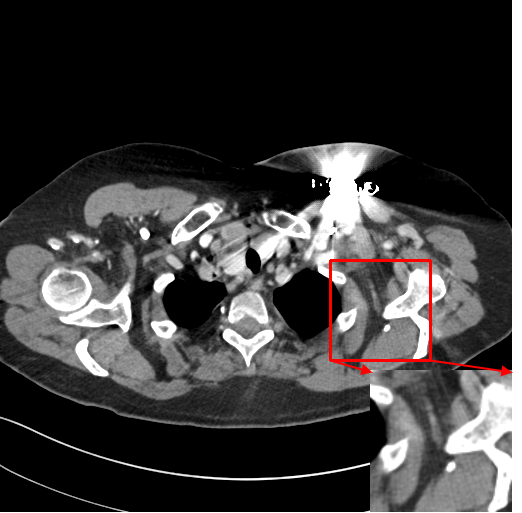}
			\centerline{\textit{\textbf{Ours}(L=2)}}
			\centerline{(\textbf{40.22/0.9832})}
		\end{minipage}
    \begin{minipage}[t]{0.245\linewidth}
			\centering
			\includegraphics[width=\textwidth]{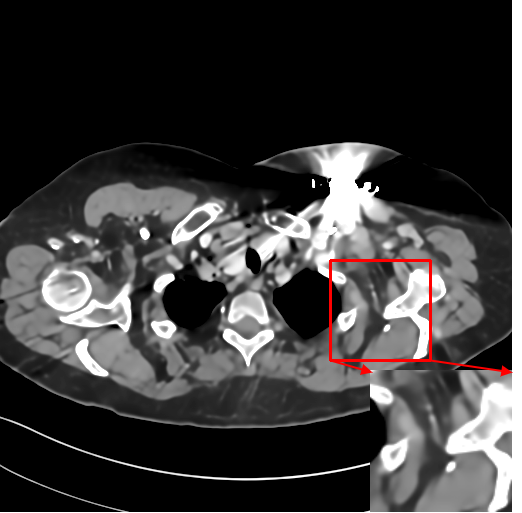}
			\centerline{InvDn \cite{Liu2021InvertibleDN}}
			\centerline{(38.20/0.9791)}
			\includegraphics[width=\textwidth]{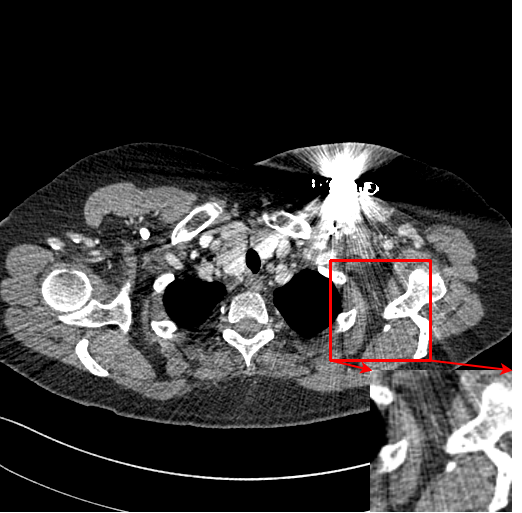}
			\centerline{NDCT}
			\centerline{(PSNR/SSIM)}
		\end{minipage}
	\caption{Visualized comparisons with state-of-the-art methods. The display window is $[160,240]$HU. Red rectangles denote ROI areas, zoomed in for better visualization.}
	\label{fig9}
\end{figure*}
\begin{table}[t]
	\caption{Quantitative results of all competing methods on Mayo.}
	\label{table5}
  \setlength{\tabcolsep}{2.7mm}{
	\begin{tabular}{l|c|c|c|c}
		\hline
		Methods &\it{\#Param (M)} & \it{MACs (G)} & \it{PSNR} & \it{SSIM} \\
		\hline
		LDCT & -- & -- & 38.13 & 0.961 \\
		BM3D \cite{Dabov2007ImageDB} & -- & -- & 42.10 & 0.983 \\
		RedCNN \cite{Chen2017LowDoseCW} & 1.85 & 462.48 & 43.36  & 0.989 \\
		WGAN-VGG \cite{Yang2018LowDoseCI} & 34.07 & 14.76 & 40.08 & 0.979\\
		\hline
		InvDn \cite{Liu2021InvertibleDN}& 2.07 & 161.6 & 42.44 &  0.987 \\
		Eformer \cite{Luthra2021EformerEE} & 1.11 & 67.45 &43.12 & 0.988 \\
		CTformer \cite{Wang2022CTformerCT} & 1.45 & 219.42 &42.76 & 0.987 \\
		\hline
		Ours(L=1) & 1.44 & 189.34 & \textbf{43.44} & \textbf{0.989}\\
		Ours(L=2) & 2.10 & 243.98 & \textbf{43.57} & \textbf{0.990}\\
		\hline
	\end{tabular}
  }
\end{table}
\subsection{Low-Dose CT Image Restoration}
\noindent We further validate our method on the medical low-dose CT image restoration. Mayo\footnote{http://www.aapm.org/GrandChallenge/LowDoseCT/}, a real clinical dataset \cite{McCollough2017LowdoseCF} authorized by Mayo Clinic for \textit{the 2016 NIH-AAPM-Mayo Clinic Low Dose CT Grand Challenge}, is used to evaluate low-dose CT (LDCT) reconstruction algorithms, i.e. recovering norm-dose CT (NDCT) image from low-dose measurements. It contains 5,936 slices with $512\times512$ sizes from 10 different subjects, each LDCT slice is simulated by inserting real noise into the NDCT to reach a noise level that corresponded to $25\%$ of the full dose. In addition, due to extra metallic applicators implanted in part patients, the heavy streak artifacts are introduced into image domain, which leads to more complex noise distribution. We shuffle the dataset and randomly select 4,000 slices as the training set, the rest are used as validation and testing to evaluate the performance of our approach. Representative methods, including BM3D, WGAN-VGG \cite{Yang2018LowDoseCI} and RedCNN \cite{Chen2017LowDoseCW}, are used for comparison. Further, to evaluate the performance of bijective mapping-based method, we select the InvDn \cite{Liu2021InvertibleDN} for comparison, where we obey the best hyperparameters setting and retrain it with $600k$ iterations. In addition, we also select recent Transformer-based methods for comprehensive comparisons, e.g., Eformer \cite{Luthra2021EformerEE} and CTformer \cite{Liu2021InvertibleDN}.

Quantitative results are illuminated in Tab.~\ref{table5}. Compared to the general CNN-based and Transformer-based methods, our method presents powerful denoising ability with a single decomposing module only, and significant gains are achieved with a 2-level decomposition, i.e. $L=2$. Our approach surpasses RedCNN by an average margin of $\sim$ 0.2dB in PSNR while with few computational costs. InvDn doesn't present obvious advantages with the bijective characteristic, where complex high-frequency distribution in CT image domain is hard to approximate with case-agnostic latent variable.

Visualized results are demonstrated in Fig.~\ref{fig9}. A representative LDCT slice with heavy streak artifacts is selected for comparison. All the methods could remove noise better. However, RedCNN could reduce the noise well, but the streak artifacts are still preserved. WGAN-VGG generates visually pleasing results with adversarial training, but it introduces extra noise and artifacts into the results. InvDn oversmoothies the local structures, meanwhile, the artifacts are also retained. Eformer and CTformer could restore global structures better, but noise and artifacts aren't removed successfully. Instead, our method has presented significant advantages in removing noise and artifacts while recovering fine structure details.

\section{Conclusion}\label{conclusion}
\noindent In this paper, we propose a flexible and efficient hierarchical disentangled representation architecture for invertible image denoising, which bridges a bijective transformation between noise image self-reconstruction and restoration, and largely mitigates the ill-posedness of the task. Extensive experiments on real image denoising, JPEG compressed artifact removal, and medical low-dose CT image restoration have demonstrated that the proposed method achieves competing performance in terms of both quantitative and qualitative evaluations, while with varying complexity. Although many concrete implementations of the advanced idea are possible, we show that a simple design already achieves excellent results on generalized image denoising tasks, which provides a new perspective for solving real image restoration tasks. 

\balance
\bibliographystyle{./IEEEtran}
\bibliography{submit}



\end{document}